\documentclass{llncs}
\usepackage{color}
\usepackage{algorithm}  
\usepackage{graphicx}
\usepackage{algpseudocode}  
\usepackage{amsmath}  

\usepackage{multirow}
\usepackage{booktabs}  
\usepackage{array} 

\newlength\savewidth
\newcommand\shline{\noalign{\global\savewidth\arrayrulewidth
                            \global\arrayrulewidth 0.8pt}%
                   \hline
                   \noalign{\global\arrayrulewidth\savewidth}}

\begin{document}
\title{Unpack Local Model Interpretation for GBDT} 
\author{Wenjing Fang \inst{1} 
\and Jun Zhou \inst{1}
\and Xiaolong Li \inst{1}
\and Kenny Q. Zhu \inst{2}
}
\institute{Ant Financial Services Group, Hangzhou, China\\  
\email{\{bean.fwj,jun.zhoujun,xl.li\}@antfin.com}
\and 
Shanghai Jiao Tong University, Shanghai, China\\ 
\email{kzhu@cs.sjtu.edu.cn}
}
\maketitle
\begin{abstract} 
A gradient boosting decision tree (GBDT), which aggregates a collection of 
single weak learners (i.e. decision trees), is widely used for data mining 
tasks. Because GBDT inherits the good performance from its ensemble essence, 
much attention has been drawn to the optimization of this model. 
With its popularization, an increasing need for model interpretation arises. 
Besides the commonly used feature importance as a global interpretation, 
feature contribution is a local measure that reveals the relationship between 
a specific instance and the related output. This work
focuses on the local interpretation and proposes an unified computation 
mechanism to get the instance-level feature contributions for GBDT in any version. 
Practicality of this mechanism is validated by the listed experiments as well as 
applications in real industry scenarios.
\end{abstract}
\section{Introduction}
Machine learning has great success in modeling data and making predictions
automatically.
In many real-world applications, we need an explanation rather than
a black-box model.
For example,  when customers apply for a loan on credit, 
the loan officers will compute their credit scores  
based on their historical behaviors. In this case, it's far from enough to 
only show the customers the final scores, and the loan officers would better
give some detailed reasons. 
While most efforts in data mining have been made on improving the accuracy 
and efficiency, which results in better models, little attention is paid to 
model interpretation for these models.
Several common measures for the variable significance have been proposed. 
Gini importance is one of the commonly used importance measure for Random Forest, 
which is derived from the Gini index\cite{breiman2001random}. 
Gini is used to measure impurity between the parent node and
two descendent nodes of samples after splitting. 
The final importance is accumulated from the Gini changes for each feature 
over all the trees in forest. 
This general feature importance(FI), also known as 
{\em global interpretation} , shows the important factors of the target, 
which unpacks the general information in the trained models. 
However, it doesn't take any feature values of an instance into consideration, 
which is insufficient sometimes. 
{\em Local interpretation}, on the other hand, places particular emphasis 
on a specific case and reveals the main causes of each record. 
This type of interpretation makes up for the shortages of the global one.
One approach proposed to define the feature contributions(FC)~\cite{palczewska2013interpreting}
 , which is accumulated from label distribution changes, as a measure of the 
feature impact on the output. The value of feature contribution reveals how 
much a feature contributes and the sign represents whether it's a positive 
impact or not. 

GBDT\cite{friedman2001greedy} is an ensemble model built on top of  
a bunch of regression decision trees. It has some appealing characteristics. 
For example, GBDT can naturally handle nonlinearity and tolerate missing values.
As a winning model in many data mining 
challenges~\cite{he2014practical,bennett2007netflix,chapelle2011yahoo}, 
GBDT is a good option for regression, classification
and ranking problems with well-known ability to generalize. 
Besides its wide range of applications, 
GBDT is also flexible in allowing users to define their
own suitable loss functions.
Furthermore, there are many 
implementations\cite{chen2016xgboost}\cite{ke2017lightgbm} 
and much work has been done to speed up the training process.

In most cases, GBDT outperforms linear models and random forest. 
Given the popularity and high quality of GBDT, it's important to uncover 
internals of the model. For GBDT, global feature importances calculation is 
widely used to do the feature selection. 
For example, Breiman proposed a method to estimate feature 
importance\cite{friedman2001greedy}. 
However, existing work has largely ignored
the exploration of local interpretations, which will be the focus
of  this paper. Specifically, we will study feature contributions for GBDT. 
We starts from previous approaches of model interpretation for random forest\cite{palczewska2013interpreting} 
and update the definition of  the feature contribution. 
The proposed mechanism is flexible enough to interpret all 
versions of GBDT. The original definition based on label distribution 
change is proved to be a special case of ours under a particular loss function. 

The rest of the paper is organized as follows. Section \ref{secrel}
provides a brief review of related work on local interpretations. Section \ref{secpre} gives out the formal definition
of feature contribution as preliminary and presents the approach for calculating feature contributions
for random forests. In section \ref{secmech}, we describe the rationale 
behind as well as main actions in interpreting GBDT. 
Section \ref{secexp} contains experiment settings and the process 
to examine the proposed methodology. At the end, section \ref{seccon} concludes 
our work.

\section{Related Work}\label{secrel}
Local model interpretation provides convincing reasons to the model outputs.
One type of interpretations prefer both the good performance of complex models and
interpretability of simple models. The pipeline of this type will first make use of
advanced models as a black-box and then extract useful information out of it with
the help of a more interpretable model. For example,
a novel approach in \cite{cui2015optimal} 
formally treats the interpretation of additive tree models as extracting 
the optimal actionable plan. It models the optimization problem as 
an integer linear programming and utilizes existing toolkit as the solver.
The constraints are based on both the output score and the objective function. 
Notice that, this kind of approaches need extra training process especially 
for the interpretation and bring new models or tasks to solve.

Some other researchers come up with model-independent local interpretations. 
They mainly make changes to feature value and test the chain effect to 
performance loss of predictions.
The loss is then taken as the measure of local importance of 
feature\cite{lei2017distribution}. This method only relies
on the output evaluation and provides an unified way to check feature 
contribution for black-box models. 
By replacing the actual
feature values with missing, zero or average values, the impact of a feature in predicting is
then removed. The instance-level contributions of all the features can be calculated separately
and compared with each other. Moreover, this method is also work for global feature importance.

As a derivative of decision tree, the random forest goes further on model interpretation
than GBDT. The method in \cite{kuz2011interpretation,palczewska2013interpreting}
computes the feature contributions so as to show informative results
about the structure of model and provide valuable information for designing new compounds.
This method makes full use of the information, not only the training data
but also the model structure. It is natural to design the interpretations with the model
structures to get a more reasonable result.

This work 
proposes an easy way to get the feature contributions on 
the instance-level. 
Generally, it can be applied to all versions of GBDT implementations
with little preprocessing and modification to the prediction process.

\section{Preliminary}\label{secpre}

Additive tree models are a powerful branch of machine learning 
but are often used as black boxes. Though they enjoy high accuracies, 
it's hard to explain  their predictions from a feature based point of view. 
Different ensemble strategies 
bring out different models while sharing the tree structure as a basis. So the model interpretations for different addictive 
tree models share some key spirits and can spread out from one to another with appropriate adaptation. In this section, 
we first review a practical interpretation method for random forest (for the binary classification) and introduce the general definition of feature
contribution to better illustrate the proposed model interpretation for GBDT. 

\subsection{Interpretation for Random Forest}\label{secrf}
Random forest is one of the most popular machine learning models due to its exordinary
accuracy utilizing categorical or numerical features on regression 
and classification problems. A random forest is a bunch of  
decision trees that are generated respectively and vote together to get a final prediction. Every tree is trained on randomly sampled
data and subsampling feature columns to introduce the diversity for better generalization, which is the key weakness of single
decision tree models. Random forest is known as a typical bagging model and the bagging strategy works out by averaging the noises to get a lower variance model. 

An instance starts a path from the root node all the way down to a leaf node according
to its real feature value. All the instances in the training data will fall into several nodes and different nodes have quite different label distributions of the 
instances in them.  Every step after passing a node, the probability of being the positive class changes with the label distributions.
All the features along the path contribute to the final prediction of a single tree.

A practical way to evaluate feature contributions is explored\cite{palczewska2013interpreting}. The key idea is taking the distribution change values for the positive class as 
the feature contribution. Concretely, it takes four procedures to work:
\begin{enumerate}
\item Computing the percentage of positive class of every node in a tree;
\item Recording the percentage difference between every parent node and its children;
\item Accumulating the contributions for every feature on each tree;
\item Averaging the feature contribution among all the trees in the forest;
\end{enumerate}

The method consists of an offline preparation embedded in training (steps 1-2) 
and an online computing with the prediction process (step 3-4). 
It is easy to record the local contribution (or local increment) and related split feature to every edge on a tree. 

\subsection{Gradient Boosting Decision Tree}
GBDT is another type of ensemble model that consists of a collection of 
regression decision trees.
However, the ensemble is based on gradient boosting which promotes the prediction gradually by reducing the residual.
For every iteration, a new model is built up to fit the negative gradient of the loss function until it converges under an acceptable threshold. 
The final prediction is the summation of all stagewise model predictions. Gradient boosting is a general framework and different models 
are available to be embedded. GBDT introduces decision tree as the basic weak learner.  When square error is chosen as the 
loss function, the residual between current prediction and target label is the negative gradient which is computational friendly.

From the above definition, we can see the differences between random forest and GBDT, some of which are the main obstacles that prevent us from adapting 
the model interpretation for random forest to GBDT:
\begin{enumerate}
\item Random forest aggregates trees by voting, while GBDT sums up the scores 
from all the trees. This means that the trees in GBDT are not equal
and the trees have to be trained in sequential order. 
The interpretation should make proper adaptations to deal with this problem.
\item Decision tree in GBDT outputs a score instead of a majority class type for classification problems. Though we can get the label 
distribution changes as random forest interpretation, the output scores in GBDT should be wisely taken into consideration.
\end{enumerate}

\subsection{Problem Statement}
Given a training dataset $D=\{x^{(i)},y^{(i)}\}_{i=1}^{N}$, where $N$ is the total number of training samples, $x=(x_{1},x_{2},...,x_{S})$ 
implies a $S$ dimensional feature vector, $x^{(i)}$ is the feature vector for the 
$i$-th sample and $y^{(i)}$  is the related label. We can 
illustrate training process of GBDT as in algorithm \ref{alg:traingbdt}. $r_{mi}$ is the residual for sample $i$ in the m-th iteration.
\begin{algorithm}[htb] 
  \begin{algorithmic}[1]  
  \caption{Gradient Boosting Decision Tree}
   \label{alg:traingbdt}  
   \Function {Train}{D,M}
   \State Init $f_{0}(x)=0$
    \For{$m=1,2,...,M$}
    \State Compute  residual:
    \State $r_{mi}=y_{i}-f_{m-1}(x_{i}),\: i=1,2,\ldots,N$
   \State Train a regression decision tree from residual:
   \State $T_{m} = $\Call{BuildTree}{$D$}
   \State Cumulated prediction sum:
   \State $f_{m}(x)=f_{m-1}(x)+T_{m}$
    \EndFor 
    \State Get finally boosting function:
    \State $f{}_{M}=\sum\limits_{m=1}\limits^{M}T_{m}$
    \State \Return $f{}_{M}$
     \EndFunction    
   \Function {PredictInstance}{$X_{i}$,$f{}_{M}$}
      \State score  = $ \sum\limits_{m=1}\limits^{M} \Call{TreePredict}{X_{i},T_{m}}$
	\State \Return score
    \EndFunction
   \end{algorithmic}  
\end{algorithm}  

Besides the basics of model, the feature contribution(FC) , as the key concept
for local interpretation, is clarified below. We introduce the notation of FC 
by denoting the model interpretation for random forest in section \ref{secrf} :
\begin{equation}  \label{li}
LI_{f}^{c}=
\left\{  
             \begin{array}{ll}  
             \multirow{2}*{$S_{c}-S_{p}$} & \quad \quad {\rm if~ the ~split~ in~ the~ parent~ is ~ performed}\ \\  & \quad \quad {\rm over~ the~ feature~} f;   \\  
             0, &  \quad\quad \rm{otherwise}
             \end{array}  
\right.  
\end{equation}  

$LI_{f}^{n}$ in equation \ref{li} is the Local Increment(LI) of feature $f$ for node $n$  defined before. For binary classification, $Y_{mean}^{n} $ 
represents the percentage of the instances belonging to the positive class in node $n$.

\begin{equation}  \label{fc1}
FC{}_{i,m}^{f}=\sum_{c\in path(i)}LI_{f}^{c}
\end{equation}  
\begin{equation}  \label{fc2}
FC_{i}^{f}=\sum_{m=1}^{M}FC_{i,m}^{f}
\end{equation}  
On a single tree $m$, $FC{}_{i,m}^{f}$ in equation \ref{fc1} cumulates the feature contribution of feature $f$ for a specific instance $i$. 
 Equation \ref{fc2} later average all the feature contribution for feature $f$ among all the trees.

\section{Mechanism}\label{secmech}
Looking back at model interpretation for random forest, its central spirit is to establish the idea of feature contribution.
By computing label distribution, a measure of the change is then obtained and associated with the split feature.  
In the case of GBDT, we can expand this computation with a slight modification. Because the targets of the latter trees
are the residual, it should replace the instance label while computing label distribution.
Nevertheless, the problem of this version is that the average of labels on a leaf node is not always equal to the score
on it. So the valuable model information in these scores are not utilized and the method is not appropriate for different GBDT 
versions \cite{friedman2001greedy,chen2016xgboost}.

In fact, the loss function determines the optimal coefficient and table \ref{tabloss} shows some common examples.
LS and LAD stand for Least Square and Least Absolute Deviation respectively. $\tilde{y}_{i}$ is the residual updated after
each iteration. $F_{m-1}(x_{i})$ is the approximation on iteration $(m-1)$. $g_{i}$ and $h_{i}$ are the first and second order
gradient statistics on the loss. Different from the numerical optimization essence to compute negative gradient (for LS and LAD), 
XGB\cite{chen2016xgboost} first approximates the loss function with its second order Taylor expansion and an analytic solution is then got.
So it contains no negative gradient computation and the evaluation of leaf weights is far from the label average. 
Particularly,  only if the LS loss function and traditional GBDT training process is used, the  label averages meet the scores.
\begin{table}[htbp]
  \centering
  \caption{Loss Functions of GBDT}
\label{tabloss}
    \begin{tabular}{llll}
     \shline
    Settings    \quad & Loss Function& \quad Negative Gradient& \quad Leaf weight \\
    \hline
     LS&    $\frac{1}{2}[y_{i}-f(x_{i})]^{2}$     &  \quad $y_{i}-f(x_{i})$   &  $\quad ave_{x_{i}\in R_{jm}}{\tilde{y}_{i}}$ \\
     \specialrule{0em}{2pt}{1pt}
     LAD&    $\mid y_{i}-f(x_{i})\mid$   &  $\quad sign[y{}_{i}-f(x_{i})]$     & \tiny $\quad median_{x_{i}\in R_{jm}}{\{y_{i}-F_{m-1}(x_{i})\}}$ \\
     \specialrule{0em}{3pt}{1pt}
   \multirow{2}*{XGB} &   \small $\sum_{i=1}^{n}[l((y_{i},\hat{y}^{(t-1)}))+g_{i}f_{t}(x_{i})$   &     \multirow{2}*{\qquad/}    &  \multirow{2}*{  $\quad -\frac{\sum_{i\in I_{j}}g_{i}}{\sum_{i\in I_{j}}h_{i}+\lambda} $}\\
  &   \small  $+\frac{1}{2}h_{i}f_{t}^{2}(x_{i}))]+\Omega(f_{t})$& & \\
   \shline
    \end{tabular}%
  \label{tab:addlabel}%
\end{table}%

\begin{figure}[htbp]
 \centering
 \includegraphics[width=0.45\textwidth]{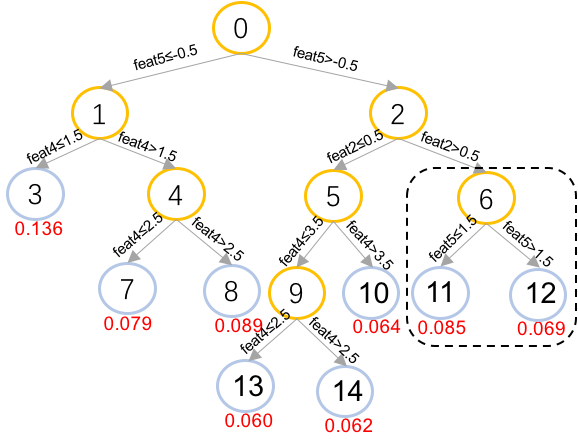}
 \includegraphics[width=0.45\textwidth]{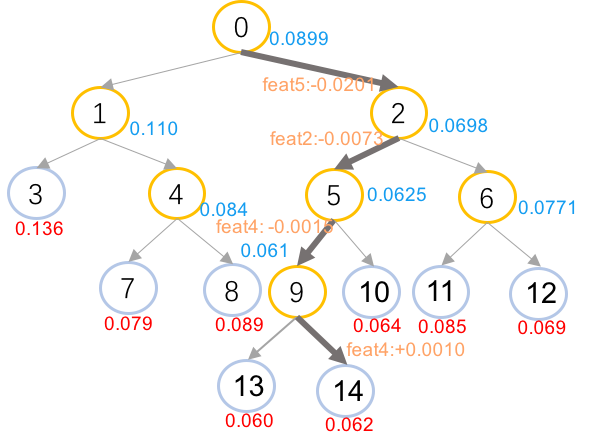}
   \caption{Feature Contribution Example for GBDT}\label{figgbdt}
\end{figure}

Without loss of generality, the interpretation for GBDT needs to work on the 
leaf scores. Since the scores are only assigned
to leaf nodes, we have to find a way to propagate them back all the way to the root. The left tree of Fig \ref{figgbdt} shows an example
tree in a GBDT model, with split feature and split value marked on arcs. Observing the three nodes in the rounded rectangle, 
the instances in node 6 will get a score difference as: $S_{n11}-S_{n12}=0.085-0.069=0.016$, where $S_{nk}$ is the score on node k. 
Moreover, this difference is caused by splitting feature $feat5$ branching by a threshold of 1.5. We can allocate this difference to the
two branches by assigning the average score of child nodes to their parent node. For instance, $S_{n6}=\frac{1}{2}(S_{n11}+S_{n12})=\frac{1}{2}\times(0.085+0.069)=0.0771$.
Then, the local increment metrics could be calculated using the scores,  $LI_{feat5}^{n11}=S_{n11}-S_{n6}=0.085-0.0771=0.0079$.
Similarly, the leaf scores as well as  the local increment  could be spread 
to the whole tree.

The interpretation process during predicting is the same as that of the random forest.  On the right hand side of Fig \ref{figgbdt},  all the node average scores and  feature contributions
on the tree are marked. Supposing an instance gets a final prediction on leaf node 14 of tree $t$, a cumulation through the path: $n0\rightarrow n2\rightarrow n5\rightarrow n9\rightarrow n14$
will be executed: $FC{}_{feat5}^{t}=LI_{feat5}^{n2}=-0.0201$, $FC{}_{feat2}^{t}=LI_{feat2}^{n5}=-0.0073$,$FC{}_{feat4}=LI_{feat4}^{n9}+LI_{feat4}^{n14}=-0.0015+0.0010=-0.0005$.

By the propagation strategy, the average score is assigned to the node 6 which assumes  an instance 
falls into the left branch or the right with equal probability. So the expectation of intermediate nodes could be revised 
as in equation \ref{avgmod}:
\begin{equation}  \label{avgmod}
S_{p}=\frac{1}{2}(S_{c1}+S_{c2})\rightarrow\frac{N_{c1}\times S_{c1}+N_{c2}\times S_{c2}}{N_{c1}+N_{c2}},
\end{equation}  
where the $N_{c1}$ and $N_{c2}$ is the number of the instances fall into child nodes node c1 and c2.
These statistics need extra information from training process.

By viewing the computation in this brand new way, 
we get a  flexible interpretation mechanism by only using the leaf
node scores and instance distributions, regardless of the implement settings of GBDT.  Under the setting of
the LS loss function, we can see that not only the label distribution meets the 
prediction score on leaf node but also the
label distribution of the intermediate node meets our back propagated score. 
That is to say, the label distribution
method is a special case of our mechanism with this particular setting. Furthermore,
this method also supports the multiple classification problems.

\section{Experiment}\label{secexp}
In this section, we demonstrate the experiments on the proposed interpretation.
In the first place,  we show the mechanism is reliable and generally agrees with global 
feature importance. Then we compare our interpretations to those of random 
forest and find it accord with the global feature importance better. Finally, 
we study the interpretations of real cases in our scenario and get a satisfied
analysis for them.

\subsection{Experiment setup}
The GBDT version in our experiment is the Scalable Multiple 
Additive Regression Tree(SMART)\cite{zhou2017psmart}, which is a distributed algorithm under the parameter server. 
Hundreds of billions of samples with thousands of features could be trained by the algorithm.
Not only the storage usage but also the running time cost is optimized 
without the loss of the accuracy.  

The training data is drawn from 
transactions under the scene of Fast Pay(FP)
in Alipay\footnote{https://global.alipay.com/}. A transaction is marked as a positive if it is reported as a fraud by the customer.
To keep a balanced ratio between positive and negative cases, only 1\% of normal transactions  are retained
 by random sampling. 

\begin{figure}[th]
 \centering
 \includegraphics[width=0.7\textwidth]{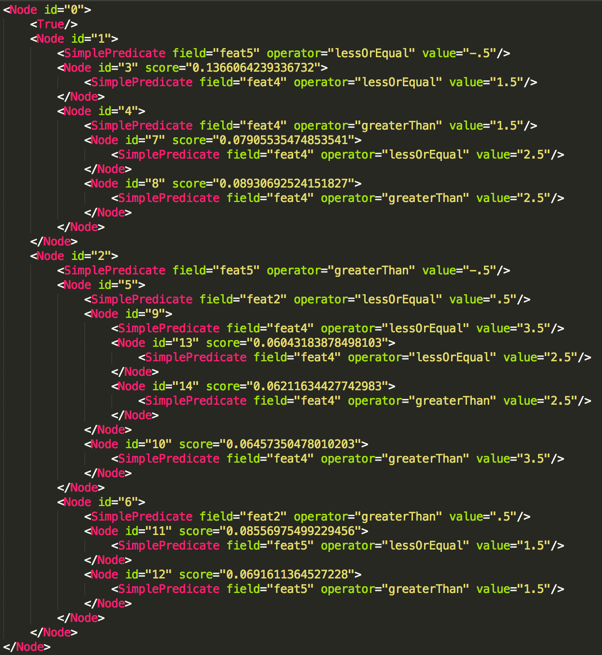}
    \caption{GBDT Model in PMML Format }\label{figpmml}
\end{figure}
Fig \ref{figpmml} is a fraction of GBDT model in Predictive Model Markup Language (PMML) format\footnote{http://dmg.org/pmml/v4-3/GeneralStructure.html}
and the tree embedded in it can be
 translate as shown in Fig \ref{figgbdt}. 
The element $Node$  is an encapsulation for a 
 tree node, which contains a predicative rule to choose itself or its siblings. The attribute $id$ 
 assigns a unique number to each node in a tree. The value of $score$ in a $Node$ is the 
 predicted value for an instance falling into it. $SimplePredicate$ is a simple boolean expression 
 indicating the split information. Our pre-trained model is stored as a PMML file. 
 JPMML\footnote{https://github.com/jpmml/jpmml-evaluator} is employed as the evaluator and
 we implement the proposed interpretation based on it.
 
\subsection{Consistency check}
We implement the feature contribution as the previous description in \cite{friedman2001greedy}. In order to make the interpretation
be independent of the training process of GBDT, the training algorithm is not changed in our experiment.
In order to get the distribution of instances in equation \ref{avgmod}, we use JPMML to predict
the training instances and record instance distributions on every node.  According to the tree structure in model and
instance distributions, the pre-process is done by back propagating the local increments  as shown in 
section \ref{secmech}. With the local increments, the feature contributions of the new instances could be computed. 
After interpreting lots of instances, we can get a distribution of feature contributions among the instances. 
The median is a robust estimator for the expectation of the general feature contribution and should somehow keep accordance with
the global feature importances metrics\cite{palczewska2013interpreting}.
\begin{figure}[htbp]
 \centering
 \includegraphics[width=0.9\textwidth]{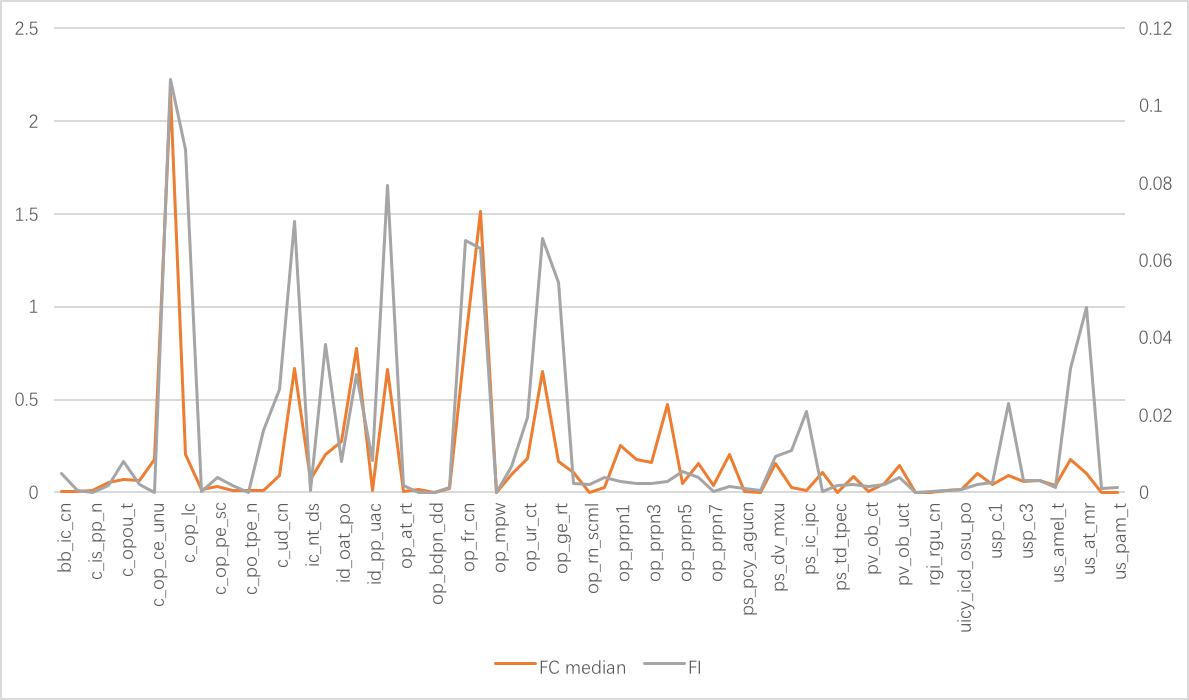}
    \caption{Feature Importance and Feature Contribution Medians}\label{figimp}
\end{figure}

Fig \ref{figimp} plots the global Feature Importance(FI) for GBDT and Feature Contribution(FC) medians for every feature.
As we can see, this two statistics have similar distributions and are in good agreement. It proves that the
interpretation for GBDT is practical and reasonable.

\subsection{Comparison to Random Forest}
Following the experiment of last section,  we get a ranking of the feature contribution median. This ranking
is a measure of feature importance and reflects the quality of local interpretation.  We implement the work for
random forest in \cite{palczewska2013interpreting} and compare it with our ranking. For justice, we replace
the GBDT Feature Importance with Information Value(IV) as the importance metric. IV is a concept from 
information theory and shows the predictive strength for the features\cite{kullback1997information}. 
\begin{figure}[htbp]
 \centering
 \includegraphics[width=0.9\textwidth]{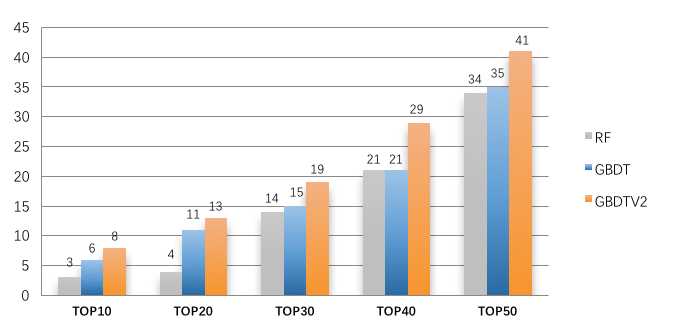}
    \caption{Interpretation: GBDT  v.s RF}\label{figiv}
\end{figure}

In Fig \ref{figiv}, we compute the intersection size on different variable coverage (i.e. Top 10-50 features
of IV).  $RF$ implies the method explained in section \ref{secrf}. $GBDT$ is the simple average strategy with
only the information in PMML file. $GBDTV2$ is the revised version in equation \ref{avgmod}. From the result,
our interpretations capture the importance better and the revised version works best.

\subsection{Case Study}
\begin{figure}[htbp]
 \centering
 \includegraphics[width=0.9\textwidth]{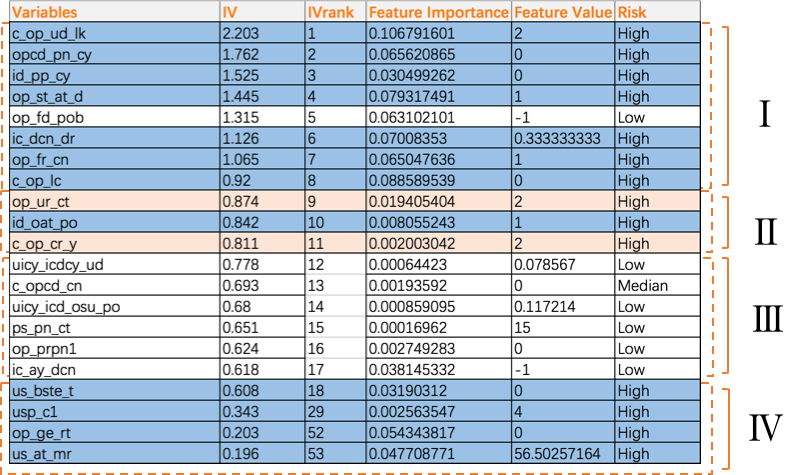}
    \caption{Case Study for Interpretation}\label{figcase}
\end{figure}

Besides the general evaluation, we analysis the 300 specific instances in the test data. Fig \ref{figcase} shows a case,
we only list some representative fields and divide them into 4 parts. The variables are ranked by IV (general feature
importance). Domain experts check the feature risk manually and draw the following conclusions:

\begin{itemize}
\item Part \uppercase\expandafter{\romannumeral1}: 
Variables in this section are with high IV, our interpretation is able to capture the features that are 
judged to be high risk(marked as blue fields). The feature with high IV but low risk (judging from
the feature value) is assigned a lower score, so the interpretation is good for instance-level contributions.
\item Part \uppercase\expandafter{\romannumeral2}:
 There are 2 variables(colored pink)  with high IV and marked high risk is missed by the interpretation, which 
 mainly due to its low occurrence in split features. The global importance of these two variables is also low and
 model interpretations are limit by the model quality.
\item Part \uppercase\expandafter{\romannumeral3}: 
Variables with median or low IVs are not caught by mistake and is assigned a low feature contribution for that case.
\item Part \uppercase\expandafter{\romannumeral4}: 
Several variables are considered to be high risk for the particular instance, even the general IVs of them are low. 
Our interpretation finds them out, which shows the superiority of the local feature contribution over the global feature importance.
\end{itemize}
Further more, if we conduct interpretations on a batch of  fraud cases which are missed by the model,
the local feature contributions will help analysts improve the model.

\section{Conclusion}\label{seccon}
Employing models as a black-box is not enough. A measure for the impact of
a feature on the prediction convinces analysts in an intuitive way. The local interpretation
provides an explanation when necessary and contributes to the promotion of the 
models. We describe a method to unpack the interpretation for the advanced model GBDT.
To the delight of analysts, the whole process is independent from the training details and 
technical optimizations. Only the tree structure and instance distribution are needed, which
can be easily extracted by a post-processing after training. The label distribution based 
method of random forest is proved to be a special case of our method. We explore the 
distribution of local feature contributions and prove it to be in agreement with global feature
importance. The method is applied to real case studies in different scenarios and serves
as a good translator of our models.  
\bibliographystyle{splncs03}
\bibliography{infocode}
\end{document}